\documentclass[twoside,11pt]{article}

\usepackage{jmlr2e}
\usepackage{amssymb}
\usepackage{amsfonts}
\usepackage{amsmath}
\usepackage{dsfont}
\usepackage{mathrsfs}
\usepackage{graphicx,float,psfrag,epsfig}
\usepackage{wrapfig}
\usepackage{verbatim}
\usepackage{rotating}

\jmlrheading{11}{2010}{2057-2078}{6/09; Revised 4/10}{7/10}{Raghunandan~H.~Keshavan, Andrea~Montanari and Sewoong~Oh}

\ShortHeadings{Matrix Completion from  Noisy Entries}{Keshavan, Montanari and Oh}

\firstpageno{2057}

\newtheorem{propo}{Proposition}[section]
\newtheorem{coro}[propo]{Corollary}
\newtheorem{thm}[propo]{Theorem}

\def\de{{\rm d}}

\def\tF{\widetilde{F}}

\def\T{{\sf P}}
\def\eps{\epsilon}

\def\reals{{\mathds R}}
\def\tM{\widetilde{M}}
\def\tZ{\widetilde{Z}}
\def\tD{\widetilde{D}}
\def\tN{\widetilde{N}}

\def\Mmax{M_{\rm max}}

\def\diag{{\rm diag}}

\def\prob{{\mathbb P}}

\def\E{\mathbb E}

\def\<{\langle}
\def\>{\rangle}
\def\diag{{\rm diag}}
\def\Grass{{\sf G}}

\def\Manif{{\sf M}}
\def\id{{\mathbf I}}
\def\cP{{\cal P}}
\def\cF{{\cal F}}

\def\Xm{{\bf x}}
\def\Wm{{\bf w}}

\def\Um{{\bf u}}
\def\Whm{\widehat{\bf w}}
\def\eps{\epsilon}
\def\Tang{{\sf T}}
\def\Trace{{\rm Tr}}
\def\hM{\widehat{M}}

\def\hW{\widehat{W}}

\def\hQ{\widehat{Q}}

\def\E{{\mathbb E}}
\def\grad{{\rm grad}\, }
\def\Co{{\cal K}}

\def\optspace{{\sc OptSpace}}
\def\ue{\underline{e}}
\def\txi{\tilde{\xi}}

\title{Matrix Completion from  Noisy Entries}

\author{\name Raghunandan~H.~Keshavan \email raghuram@stanford.edu \\
       \name Andrea~Montanari\thanks{Also in Department of Statistics.} \email montanari@stanford.edu \\
       \name Sewoong~Oh \email swoh@stanford.edu \\
       \addr Department of Electrical Engineering\\
       Stanford University\\
       Stanford, CA 94304, USA
	}

\editor{Tommi Jaakkola}

\begin{document}
\maketitle

\begin{abstract}%
Given a matrix $M$ of low-rank, we consider the problem of
reconstructing it from noisy observations of a small,
random subset of its entries. The problem arises in a variety
of applications, from collaborative filtering (the `Netflix problem')
to structure-from-motion and positioning. We study a low complexity algorithm
introduced by \cite*{KOM09}, based on a combination
of spectral techniques and manifold optimization, that we call 
here {\sc OptSpace}. We prove performance guarantees that are 
order-optimal in a number of circumstances.
\end{abstract}


\section{Introduction}

Spectral techniques are an authentic workhorse in machine learning,
statistics, numerical analysis, and signal processing.  Given 
a matrix $M$, its largest singular values---and the associated singular
vectors---`explain' the most significant correlations in the 
underlying data source. A low-rank approximation of $M$ can further
be used for low-complexity implementations of a number of
linear algebra algorithms \citep{FKV}.

In many practical circumstances we have access only to a sparse subset of
the entries of an $m\times n$ matrix $M$. It has  
recently been discovered that, if the matrix $M$ has
rank $r$, and unless it is too `structured', a small random subset of its 
entries allow to reconstruct it \emph{exactly}.
This result was first proved by \cite{CaR08}
by analyzing a convex relaxation introduced by \cite{Fazel}.
A tighter analysis of the same convex relaxation was 
carried out by \cite{CandesTaoMatrix}. A number of iterative
schemes to solve the convex optimization problem appeared soon thereafter
\citep{CCS08,FPCA,APG}.

In an alternative line  of work, \citet*{KOM09}
attacked the same problem using a combination of spectral techniques
and manifold optimization: We will refer to their algorithm as
 \optspace. \optspace\, is intrinsically of low complexity,  
the most complex operation being computing $r$ singular values (and the corresponding singular vectors)  
of a sparse $m\times n$ matrix.  
The performance guarantees proved by \cite{KOM09} are comparable 
with the information theoretic lower bound: roughly 
$nr\max\{r,\log n\}$ random entries are needed
to reconstruct $M$ exactly (here we assume $m$ of order $n$).
A related approach was also developed
by \cite{ADMiRA}, although without performance guarantees for matrix 
completion.

The above results crucially rely on the assumption that 
$M$ is \emph{exactly} a rank $r$ matrix. For many applications of interest,
this assumption is unrealistic and it is therefore important
to investigate their robustness. Can the above approaches be generalized
when the underlying data is `well approximated' by a rank
$r$ matrix? This question was addressed by \cite{CandesPlan}
within the convex relaxation approach of \cite{CaR08}.
The present paper proves a similar robustness result for \optspace. 
Remarkably the guarantees we obtain are
order-optimal in a variety of circumstances, and improve over the
analogous results of \cite{CandesPlan}.
%
%
\subsection{Model Definition}\label{sec:model}

Let $M$ be an $m\times n$ matrix of rank $r$, that is
\begin{eqnarray}
  M = U\Sigma V^T\, .\label{eq:MatrixForm}
\end{eqnarray}
where $U$ has dimensions $m \times r$, 
$V$ has dimensions $n\times r$, and $\Sigma$ is a diagonal 
$r\times r$ matrix.
We assume that each entry of $M$ is perturbed, thus producing
an `approximately' low-rank matrix $N$, with 
\begin{eqnarray*}
N_{ij} = M_{ij}+Z_{ij}\, ,
\end{eqnarray*}
where the matrix $Z$ will be assumed to be `small' in an appropriate sense.

Out of the $m\times n$ entries of $N$, a subset 
$E\subseteq[m]\times [n]$ is revealed.
We let  $N^E$ be the $m\times n$ matrix that contains the
revealed entries of $N$, and is
filled with $0$'s in the other positions
 \begin{eqnarray*}
  N^E_{ij} = \left\{
              \begin{array}{rl}
              N_{ij} & \text{if } (i,j)\in E\, ,\\
              0       & \text{otherwise.}
              \end{array} \right.
  \end{eqnarray*}
Analogously, we let $M^E$ and $Z^E$ be the $m \times n$ matrices that contain 
the entries of $M$ and $Z$, respectively, 
in the revealed positions and is filled with $0$'s in the other positions. 
The set $E$ will be uniformly  random given its 
size $|E|$.
%
%
\subsection{Algorithm}\label{sec:algorithm}

For the reader's convenience, we recall the
algorithm  
introduced by \cite{KOM09}, which we will analyze here.
The basic idea is to minimize the cost function
$F(X,Y)$, defined by
\begin{eqnarray}
F(X,Y) & \equiv &\min_{S\in \reals^{r\times r}}\cF(X,Y,S)\, ,\label{eq:MinimizeS}\\
\cF(X,Y,S)&\equiv &  \frac{1}{2}\sum_{(i,j)\in E}(N_{ij}-(XSY^T)_{ij})^2 \nonumber \, .
\end{eqnarray}
Here $X\in \reals^{n\times r}$, $Y\in\reals^{m\times r}$ are 
orthogonal matrices, normalized by $X^{T}X = m\id$, $Y^TY = n\id$.

Minimizing $F(X,Y)$ is an \emph{a priori} difficult task, since 
$F$ is a non-convex function. The key insight is that the singular 
value decomposition (SVD) of $N^E$ provides an excellent initial guess,
and that the minimum can be found with high probability by
standard gradient descent after this initialization.
Two caveats must be added to this description: 
$(1)$ In general the matrix $N^E$ must be `trimmed' to eliminate
over-represented rows and columns; $(2)$ For technical reasons,
we consider a slightly modified cost function to be denoted by
$\tF(X,Y)$.

\phantom{a}

\begin{tabular}{ll}
\hline
\vspace{-0.43cm}\\
\multicolumn{2}{l}{ {\sc OptSpace}( matrix $N^E$ )}\\
\hline
\vspace{-0.4cm}\\
1: & Trim $N^E$, and let $\tN^E$ be the output;\\
2: & Compute the rank-$r$ 
projection of $\tN^E$, $\T_r(\tN^E)=X_0S_0Y_0^T$;\\
3: & Minimize $\tF(X,Y)$ through gradient descent,
with initial condition $(X_0,Y_0)$.\\
\hline
\end{tabular}

\phantom{a}

We may note here that the rank of the matrix $M$, if not known, can be 
reliably estimated from $\tN^E$ \citep{KOImpl09}. 

The various steps of the above algorithm are defined as follows.

{\sf Trimming}. We say that a row is `over-represented' if it
contains more than $2|E|/m$ revealed entries (i.e., more than twice the 
average number of revealed entries per row).
Analogously, a column is over-represented if it contains more than 
$2|E|/n$ revealed entries. The trimmed matrix $\tN^E$ is obtained from $N^E$
by setting to $0$ over-represented rows and columns.

{\sf Rank-$r$ projection}. Let 
\begin{eqnarray*}
\tN^E = \sum_{i=1}^{\min(m,n)}\sigma_i  x_iy_i^T\, ,
\end{eqnarray*}
be the singular value decomposition of $\tN^E$,
with singular values $\sigma_1\ge\sigma_2\ge\dots$. We then define
\begin{eqnarray*}
\T_r(\tN^E) =  
\frac{mn}{|E|}\,\sum_{i=1}^{r}\sigma_i  x_iy_i^T\, .
\end{eqnarray*}
Apart from an overall normalization, $\T_r(\tN^E)$ is the best
rank-$r$ approximation to $\tN^E$ in Frobenius norm.

{\sf Minimization}. The modified cost function $\tF$ is defined
as
\begin{eqnarray}
\tF(X,Y) &=& F(X,Y) + \rho \, G(X,Y) \nonumber \\
&\equiv& F(X,Y) + \rho\sum_{i=1}^mG_1\left(\frac{\|X^{(i)}\|^2}{3\mu_0r}
\right)+
\rho\sum_{j=1}^nG_1\left(\frac{\|Y^{(j)}\|^2}{3\mu_0r}\right)\, ,
\nonumber
\end{eqnarray}
where $X^{(i)}$ denotes the $i$-th row of $X$, and
$Y^{(j)}$ the $j$-th row of $Y$. The function $G_1:\reals^+\to \reals$
is such that $G_1(z) = 0$ if $z\le 1$  and $G_1(z)=e^{(z-1)^2}-1$
otherwise. Further, we can choose $\rho = \Theta(|E|)$.

Let us stress that the regularization term is mainly introduced for 
our proof technique to work (and a broad family of functions $G_1$
would work as well). In numerical experiments we did not find
any performance loss in setting $\rho=0$.

One important feature  of \optspace\, is that 
$F(X,Y)$ and $\tF(X,Y)$ are regarded as functions
of the $r$-dimensional subspaces of $\reals^m$ and $\reals^n$
generated (respectively) by the columns of $X$ and $Y$.
This interpretation is justified by the fact that
$F(X,Y) = F(XA,YB)$ for any two orthogonal matrices 
$A$, $B\in \reals^{r\times r}$ (the same property holds for $\tF$).
The set of $r$ dimensional subspaces of $\reals^m$ is 
a differentiable Riemannian manifold $\Grass(m,r)$ (the Grassmann
manifold).  
The gradient descent algorithm is applied to the function
$\tF:\Manif(m,n) \equiv \Grass(m,r)\times\Grass(n,r)\to \reals$.
For further details on optimization by gradient descent on matrix manifolds we
refer to \cite{Edelman} and \cite{ManifBook}.
%
%
\subsection{Some Notations}\label{sec:coherence}

The matrix $M$ to be reconstructed takes the form 
(\ref{eq:MatrixForm}) where $U\in\reals^{m\times r}$,
$V\in\reals^{n\times r}$. We 
write $U=[u_1,u_2,\dots,u_r]$ and $V= [v_1,v_2,\dots,v_r]$
for the columns of the two factors, with $\|u_i\|=\sqrt{m}$, 
$\|v_i\| = \sqrt{n}$, and $u_i^Tu_j=0$, $v_i^Tv_j=0$ for $i\neq j$
(there is no loss of generality in this, since normalizations
 can be absorbed by redefining $\Sigma$).

We shall write $\Sigma = \diag(\Sigma_1,\dots,\Sigma_r)$
with $\Sigma_1\ge \Sigma_2\ge\cdots \ge\Sigma_r > 0$.
The maximum and minimum singular values will also be denoted by 
$\Sigma_{\rm max} = \Sigma_1$ and $\Sigma_{\rm min}= \Sigma_r$.
Further, the maximum size of an entry of $M$ is
$M_{\rm max} \equiv \max_{ij}|M_{ij}|$.

Probability is taken with respect to the uniformly random 
subset $E\subseteq [m]\times [n]$ given $|E|$ and 
(eventually) the noise matrix $Z$.
Define $\eps\equiv|E|/\sqrt{mn}$. In the case when $m=n$, 
$\eps$ corresponds to the average number of revealed entries per row or column.
Then it is convenient to work with a model in which 
each entry is revealed independently with probability
$\eps/\sqrt{mn}$. Since, with high probability
$|E|\in [\eps\sqrt{\alpha}\, n- A\sqrt{n\log n},\eps\sqrt{\alpha}\, 
n+ A\sqrt{n\log n}]$, any guarantee on the algorithm performances that holds
within one model, holds within the other model as well
if we allow for a vanishing shift in $\eps$.
We will use $C$, $C'$ etc.
to denote universal numerical constants.

It is convenient to define the following projection operator $\cP_E(\cdot)$ as the sampling operator, 
which maps an $m\times n$ matrix onto an $|E|$-dimensional subspace in $\reals^{m \times n}$
 \begin{eqnarray}
  \cP_E(N)_{ij} = \left\{
              \begin{array}{rl}
              N_{ij} & \text{if } (i,j)\in E\, ,\\
              0       & \text{otherwise.}
              \end{array} \right. \nonumber
  \end{eqnarray}

Given a vector $x\in\reals^n$, $\|x\|$ will denote its Euclidean norm. 
For a matrix $X\in\reals^{n\times n'}$, $\|X\|_F$ is its Frobenius norm,
and $\|X\|_2$ its operator norm (i.e., $\|X\|_2= \sup_{u\neq 0}\|Xu\|/\|u\|$). 
The standard scalar product between vectors or matrices will sometimes
be indicated by $\<x,y\>$ or $\<X,Y\> \equiv\Trace(X^TY)$, respectively.
Finally, we use the standard combinatorics notation
$[n]= \{1,2,\dots,n\}$ to denote the set of first $n$ integers.
%
%
\subsection{Main Results}\label{sec:MainResults}

Our main result is a performance guarantee for {\sc OptSpace}
under appropriate incoherence assumptions, and is
presented in Section \ref{sec:MainOptSpace}. Before presenting
it, we state a theorem of independent interest that 
provides an error bound on the simple trimming-plus-SVD approach. 
The reader interested in the {\sc OptSpace} guarantee
can go directly to Section \ref{sec:MainOptSpace}.

Throughout this paper, without loss of generality, 
we assume $\alpha\equiv m/n \geq 1$.

\subsubsection{Simple SVD}

Our first result shows that, in great generality, the 
rank-$r$ projection of $\tN^E$ provides a reasonable approximation of $M$. 
We define $\tZ^E$ to be an $m \times n$ matrix obtained from $Z^E$, 
after the trimming step of the pseudocode above, that is,
by setting to zero the over-represented rows and columns.
\begin{thm} \label{thm:main1}
Let $N= M+Z$, where $M$ has rank $r$, and 
assume that the subset of revealed entries $E\subseteq [m]\times [n]$
is uniformly random with size $|E|$. Let $\Mmax = \max_{(i,j) \in [m] \times [n]} |M_{ij}|$. Then there exists numerical constants 
$C$ and $C'$ such that
\begin{eqnarray*}
\frac{1}{\sqrt{mn}}\|M-\T_r(\tN^E)\|_F\le C \Mmax\, \left(\frac{nr \alpha^{3/2} }{|E|}\right)^{1/2}\, +\, C'  \frac{n\sqrt{r\alpha}}{|E|}\, \|\tZ^E\|_2\, ,
\end{eqnarray*}
with probability larger than $1-1/n^3$.
\end{thm}
Projection onto rank-$r$ matrices through SVD is a pretty standard
tool, and is used as first analysis method for many practical problems. 
At a high-level, projection onto rank-$r$ matrices can be interpreted
as `treat missing entries as zeros'.
This theorem shows that this approach is reasonably robust
if the number of observed entries is as large as the number of degrees of 
freedom (which is about $(m+n)r$) times a large constant.
The error bound is the sum of two contributions: the first one can be 
interpreted as an undersampling effect (error induced by missing entries)
and the second as a noise effect.
Let us stress that trimming is crucial for achieving this guarantee.
%
%
\subsubsection{OptSpace}\label{sec:MainOptSpace}

Theorem \ref{thm:main1} helps to set the stage for the 
key point of this paper:
\emph{a much better approximation is obtained by 
minimizing the cost $\tF(X,Y)$ (step 3 in the pseudocode above), provided $M$
satisfies an appropriate incoherence condition.}
Let $M = U\Sigma V^T$ be a low rank matrix, and assume, without 
loss of generality, $U^TU = m\id$ and $V^TV = n\id$. We say that $M$ is
$(\mu_0,\mu_1)$-\emph{incoherent} if the following conditions hold.
\begin{itemize}
\item[{\bf A1.}] For all  $i\in [m]$, $j\in [n]$ we have,  
           $\sum_{k=1}^{r}{U_{ik}^2} \le \mu_0 r$,
           $\sum_{k=1}^{r}{V_{ik}^2} \le \mu_0 r $.
%
\item[{\bf A2.}] For all $i\in [m]$, $j\in [n]$ we have,  $|\sum_{k=1}^{r}{U_{ik}
(\Sigma_k/\Sigma_1)V_{jk}}|\leq\mu_1r^{1/2}$.
\end{itemize}
\begin{thm} \label{thm:main2}
Let $N = M+Z$, where $M$ is a $(\mu_0,\mu_1)$-incoherent matrix
of rank $r$, and 
assume that the subset of revealed entries $E\subseteq [m]\times [n]$
is uniformly random with size $|E|$. 
Further, let $\Sigma_{\rm min}= \Sigma_r\le\dots\le\Sigma_{1}=
\Sigma_{\rm max}$ with $\Sigma_{\rm max}/\Sigma_{\rm min}\equiv \kappa$.
Let $\hM$ be the output of \optspace\,  on input $N^E$. 
Then there exists numerical constants 
$C$ and $C'$ such that if
\begin{eqnarray*}
|E| & \ge & C n\sqrt{\alpha}\kappa^2\,
\max\left\{ \mu_0 r\sqrt{\alpha}\log n\,;\,\mu_0^2r^2\alpha\kappa^4\,;\,\mu_1^2 r^2 \alpha \kappa^4 \right\}\ ,
\end{eqnarray*}
then, with probability at least $1-1/n^3$,
\begin{eqnarray}
\frac{1}{\sqrt{mn}}\, \|\hM-M\|_F\le C'\,
\kappa^2\frac{n\sqrt{r \alpha}}{|E|}\|Z^{E}\|_2\, .\label{eq:MainBound}
\end{eqnarray}
provided that the right-hand side is smaller than $\Sigma_{\rm min}$.
\end{thm}

As discussed in the next section, this
theorem captures rather sharply the effect of important classes
of noise on the performance of {\sc OptSpace}.
%
%
\subsection{Noise Models}
\label{sec:NoiseModels}

In order to make sense of the above results, it is convenient 
to consider a couple of simple models for the noise matrix 
$Z$:\vspace{0.2cm}

\emph{Independent entries model.} We assume that $Z$'s entries
are i.i.d. random variables, with zero mean $\E\{Z_{ij}\}=0$ and 
sub-Gaussian tails. The latter means that
\begin{eqnarray*}
\prob\{|Z_{ij}|\ge x\}\le 2\, e^{-\frac{x^2}{2\sigma^2}}\, ,
\end{eqnarray*}
for some constant $\sigma^2$ uniformly bounded in $n$.

\emph{Worst case model.} In this model $Z$ is arbitrary, but we 
have an uniform bound on the size of its entries: $|Z_{ij}|\le Z_{\rm max}$.
\vspace{0.2cm}

The basic parameter entering our main results is the operator
norm of $\tZ^E$, which is bounded as follows in these two noise models.
\begin{thm}
\label{thm:noisy}
If $Z$ is a random matrix drawn according to 
the independent entries model, 
then for any sample size $|E|$ 
there is a constant $C$ such that,
\begin{eqnarray}
%
\|\tZ^E\|_2\le C \sigma \left({\frac{|E|\log n}{n}}\right)^{1/2}\, ,
\label{eq:boundindepcase}
\end{eqnarray}
with probability at least $1-{1}/{n^3}$. 
Further there exists a constant $C'$ such that, 
if the sample size is $|E|\ge n\log n$
(for $n\ge \alpha$), we have
\begin{eqnarray}
%
\|\tZ^E\|_2\le C' \sigma \left({\frac{|E|}{n}}\right)^{1/2}\, ,
\label{eq:boundindepcase2}
\end{eqnarray}
with probability at least $1-{1}/{n^3}$.

If $Z$ is a matrix from the worst case model, then 
\begin{eqnarray*}
\|\tZ^E\|_2 \le 
\frac{2|E|}{n\sqrt\alpha}\,
Z_{\rm max}\, ,
\end{eqnarray*}
for any realization of $E$.
\end{thm}
It is elementary to show that, if $|E|\ge 15\alpha n\log n$,
no row or column is over-represented with high probability.
It follows that in the regime of $|E|$ for which the conditions of 
Theorem \ref{thm:main2} are satisfied, 
we have $Z^E=\tZ^E$ and hence the bound (\ref{eq:boundindepcase2})
applies to $\|\tZ^E\|_2$ as well.
Then, among the other things, this result implies that 
for the independent entries model
the right-hand side of our error estimate, Eq.~(\ref{eq:MainBound}), 
is with high probability smaller than $\Sigma_{\rm min}$, 
if $|E| \geq Cr\alpha n \, \kappa^4(\sigma/\Sigma_{\rm min})^2$.
For the worst case model, 
the same statement is true 
if $Z_{\rm max}\leq{\Sigma_{\rm min}}/{C\sqrt{r}\kappa^2} $.

%
%
\subsection{Comparison with Other Approaches to Matrix Completion}

\begin{figure}
\begin{center}
\includegraphics[width=10.cm]{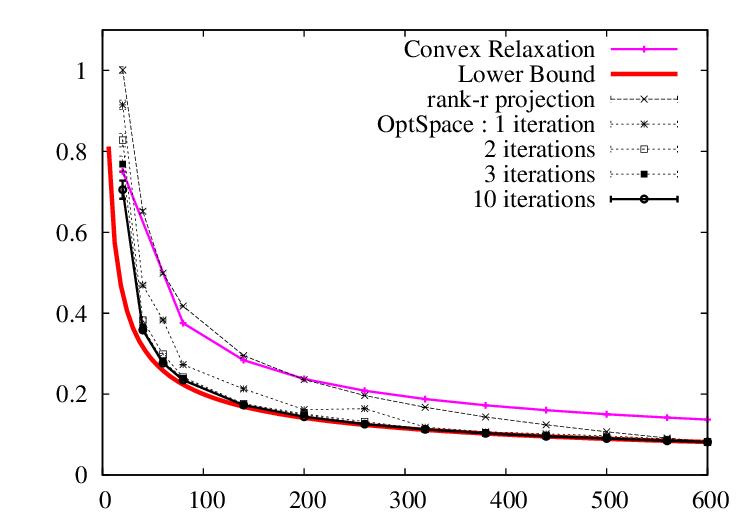}
\put(-140,-10){\small{$|E|/n$}}
\put(-278,83){\begin{sideways}\small{RMSE}\end{sideways}}
\end{center}
\caption{{Numerical simulation with random rank-$2$ $600 \times 600$ matrices.
Root mean square error achieved by \optspace\, is shown as a function of 
the number of observed entries $|E|$ and of the number of line 
minimizations. The performance of nuclear norm minimization 
and an information theoretic lower bound are also shown.} }
\label{fig:Comparison}
\end{figure}

\begin{figure}
\begin{center}
\includegraphics[width=10.cm]{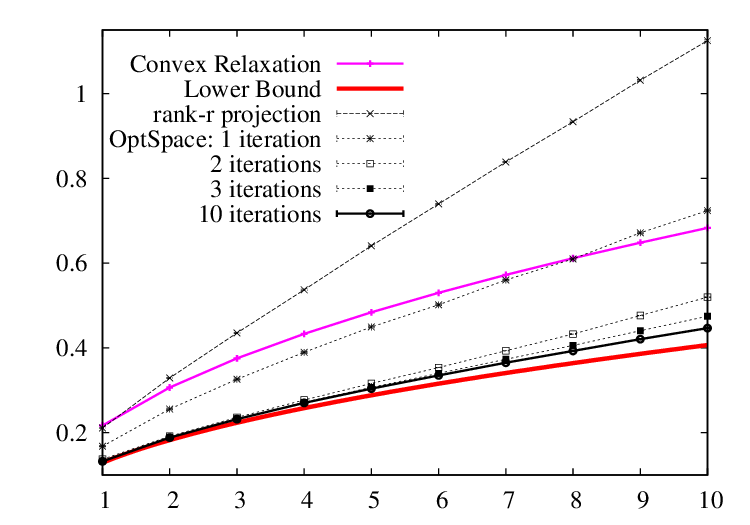}
\put(-141,-10){\small{Rank}}
\put(-278,83){\begin{sideways}\small{RMSE}\end{sideways}}
\end{center}
\caption{{Numerical simulation with random rank-$r$ $600 \times 600$ matrices 
and number of observed entries $|E|/n=120$.
Root mean square error achieved by \optspace\, 
is shown as a function of the rank and of the number of line minimizations. 
The performance of nuclear norm minimization 
and an information theoretic lower bound are also shown.} }
\label{fig:Comparison2}
\end{figure}

Let us begin by mentioning that a statement
analogous to our preliminary Theorem \ref{thm:main1} was proved by
\cite{AchlioptasRank}. Our result however applies to any number
of revealed entries, while the one of \cite{AchlioptasRank} requires
$|E|\ge (8\log n)^4n$ (which for $n\le 5\cdot 10^8$ is larger than $n^2$).
We refer to Section \ref{sec:trimming} for further discussion of this point. 

As for Theorem \ref{thm:main2}, we will mainly compare our algorithm with 
the convex relaxation approach recently analyzed by \cite{CandesPlan},
and based on semidefinite programming.
Our basic setting is indeed the same, while  the algorithms are 
rather different.

Figures \ref{fig:Comparison} and \ref{fig:Comparison2} compare the average root mean
square error $\|\hM-M\|_F/\sqrt{mn}$ for the two algorithms as a function of $|E|$ and the rank-$r$ respectively.
Here $M$ is a random rank $r$ matrix of dimension 
$m=n=600$, generated by letting $M=\widetilde{U}\widetilde{V}^T$
with $\widetilde{U}_{ij},\widetilde{V}_{ij}$ i.i.d. $N(0,20/\sqrt{n})$.
The noise is distributed according to the independent noise
model with $Z_{ij}\sim N(0,1)$. 
In the first suite of simulations, presented  
in Figure \ref{fig:Comparison}, the rank is fixed to $r=2$.  
In the second one (Figure \ref{fig:Comparison2}), 
the number of samples is fixed to $|E|=72000$. 
These examples are taken from
\citet[Figure 2]{CandesPlan}, from which we took 
the data points for the convex relaxation approach,
as well as the  information theoretic lower bound 
described  later in this section. 
After a few iterations, \optspace\, has a smaller root mean square
error than the one produced by convex relaxation.
In about 10 iterations it becomes 
indistinguishable from the information theoretic lower bound for small ranks. 

\begin{figure}
\begin{center}
\includegraphics[width=10.cm]{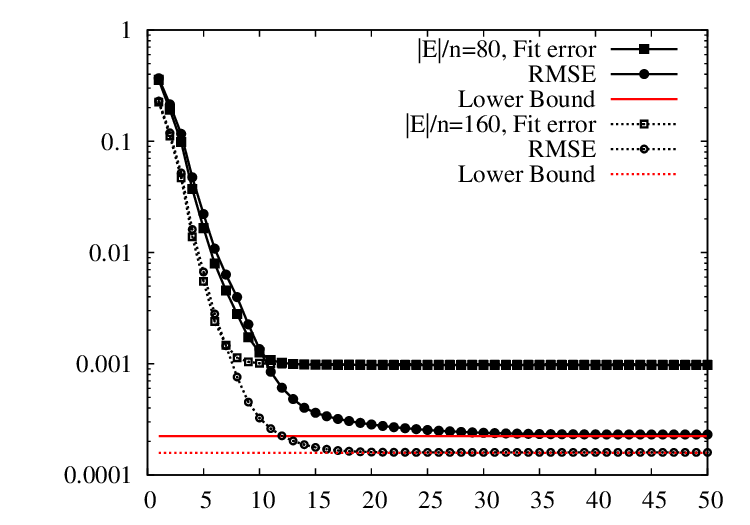}
\put(-142,-9){\small{Iterations}}
\put(-267,90){\begin{sideways}\small{Error}\end{sideways}}
\end{center}
\caption{{Numerical simulation with random rank-$2$ $600 \times 600$ matrices 
and number of observed entries $|E|/n=80$ and $160$. 
The standard deviation of the i.i.d. Gaussian noise is $0.001$. 
Fit error and root mean square error achieved by \optspace\, 
are shown as functions of the number of line minimizations. 
Information theoretic lower bounds are also shown.} }
\label{fig:Convergence}
\end{figure}

In Figure \ref{fig:Convergence}, we illustrate the rate of convergence of {\sc OptSpace}. 
Two metrics, root mean squared error(RMSE) and fit error 
$\|\cP_E(\hM-N)\|_F/\sqrt{|E|}$, are shown as functions  
of the number of iterations in the manifold optimization step. 
Note, that the fit error can be easily evaluated since $N^E=\cP_E(N)$ 
is always available at the estimator.
$M$ is a random $600\times 600$ rank-$2$ matrix generated as in the previous examples. 
The additive noise is distributed as $Z_{ij}\sim N(0,\sigma^2)$ with $\sigma = 0.001$ 
(A small noise level was used in order to trace the RMSE evolution over
many iterations).
Each point in the figure is the averaged over $20$ random instances,
and resulting errors for two different values of 
sample size $|E|=80$ and $|E|=160$ are shown. 
In both cases, we can see that the RMSE 
converges to the information theoretic lower bound 
described later in this section. 
The fit error decays exponentially with 
the number iterations and converges to 
the standard deviation of the noise which is $0.001$.
This is a lower bound on the fit error when $r \ll n$, since 
even if we have a perfect reconstruction of $M$, 
the average fit error is still $0.001$.

For a more complete numerical comparison between various 
algorithms for matrix completion, including different noise
models, real data sets and ill conditioned matrices, 
we refer to \cite{KOImpl09}.

Next, let us compare our main result with the performance guarantee
of \citet[Theorem 7]{CandesPlan}. Let us stress that we require
the condition number $\kappa$ to be bounded,
while the analysis of \cite{CandesPlan} and \cite{CandesTaoMatrix} requires a 
stronger incoherence assumption (compared to our {\bf A1}). 
Therefore the assumptions are not directly comparable.
As far as the error bound is concerned,
\cite{CandesPlan} proved that the semidefinite programming approach
returns an estimate $\hM$ which satisfies
\begin{eqnarray}
\frac{1}{\sqrt{mn}}\, \|\hM_{\rm SDP}-M\|_F\le 7 \, \sqrt{\frac{n}{|E|}}\, \|Z^E\|_F
+\frac{2}{n\sqrt{\alpha}}\, \|Z^E\|_F\, .\label{eq:CandesPlan}
\end{eqnarray}
(The constant in front of the first term
is in fact slightly smaller than $7$ in \cite{CandesPlan}, but in any 
case larger than $4\sqrt{2}$. We choose to quote a result which
is slightly less accurate but easier to parse.) 

Theorem \ref{thm:main2} improves over this result in several respects:
$(1)$ We do not have the second term on the right-hand side of
(\ref{eq:CandesPlan}), that actually increases with the number of observed entries; $(2)$ Our error decreases as $n/|E|$ rather than $(n/|E|)^{1/2}$;
$(3)$ The noise enters Theorem \ref{thm:main2} through the operator norm
$\|Z^E\|_2$ instead of its Frobenius norm $\|Z^E\|_F\ge \|Z^E\|_2$.
For $E$ uniformly random, one expects $\|Z^E\|_F$ to be roughly of
order $\|Z^E\|_2\sqrt{n}$.
For instance, within the independent entries model with bounded variance 
$\sigma$, $\|Z^E\|_F =\Theta(\sqrt{|E|})$ while $\|Z^E\|_2$ is of order 
$\sqrt{|E|/n}$ (up to logarithmic terms).

Theorem \ref{thm:main2} can also be compared to an 
information theoretic lower bound computed by \cite{CandesPlan}. 
Suppose, for simplicity, $m=n$ and assume that an oracle 
provides us a linear subspace $T$ where the correct rank $r$
matrix $M=U\Sigma V^T$ lies. More precisely, we know that $M\in T$ where
$T$ is a linear space of dimension $2nr-r^2$ defined by
\begin{eqnarray*}
 T = \{UY^T+XV^T\;|\;X\in\reals^{n\times r},Y\in\reals^{n\times r}\}\;.
\end{eqnarray*}
Notice that the rank constraint is therefore replaced by this simple linear 
constraint. 
The minimum mean square error estimator is computed by 
projecting the revealed entries onto the subspace $T$,
which can be done by solving a least squares problem.
\cite{CandesPlan} analyzed the 
root mean squared error of the resulting  estimator $\hM$  
and showed that 
\begin{eqnarray*}
\frac{1}{\sqrt{mn}}\, \|\hM_{\rm Oracle}-M\|_F \approx \sqrt{\frac{1}{|E|}}\, \|Z^E\|_F\, .
\end{eqnarray*}
Here $\approx$ indicates that the  root mean squared error 
concentrates in probability around the right-hand side. 

For the sake of
comparison, suppose we have i.i.d. Gaussian noise with variance $\sigma^2$.
In this case the oracle estimator yields (for $r = o(n)$)
\begin{eqnarray*}
\frac{1}{\sqrt{mn}}\, \|\hM_{\rm Oracle}-M\|_F \approx 
\sigma\sqrt{\frac{2nr}{|E|}}\, .
\end{eqnarray*}
The bound (\ref{eq:CandesPlan}) on the semidefinite programming approach
yields
\begin{eqnarray*}
\frac{1}{\sqrt{mn}}\, \|\hM_{\rm SDP}-M\|_F\le \sigma\,\Big(7 \, \sqrt{n|E|}
+\frac{2}{n}|E|\Big)\, .
\end{eqnarray*}
Finally, using Theorems \ref{thm:main2} and \ref{thm:noisy}
we deduce  that {\sc OptSpace} achieves 
\begin{eqnarray*}
\frac{1}{\sqrt{mn}}\, \|\hM_{\rm OptSpace}-M\|_F\le 
\sigma\,\sqrt{\frac{ C\,nr}{|E|}}\, .
\end{eqnarray*}
Hence, when the noise is i.i.d. Gaussian with small enough $\sigma$, 
{\sc OptSpace} is order-optimal.
%
%
\subsection{Related Work on Gradient Descent}

Local optimization techniques such as gradient descent 
of coordinate descent have been intensively studied in machine learning,
with a number of applications. Here we will briefly 
review the recent literature on the use of such techniques within 
collaborative filtering applications. 

Collaborative filtering was studied from a graphical models
perspective in \cite{SMH07}, which introduced an approach
to prediction based on Restricted Boltzmann Machines (RBM).
Exact learning of the model parameters is intractable 
for such models, but the authors studied the performances 
of a \emph{contrastive divergence}, which computes an approximate 
gradient of the likelihood function, and uses it to optimize the 
likelihood locally.
Based on empirical evidence, it was argued that RBM's have several advantages
over spectral methods for collaborative filtering.

An objective function analogous to the one used in the present paper was
considered early on in \cite{SJ03}, which uses gradient descent in 
the factors to minimize a weighted sum of square residuals. 
\cite{SM08} justified the use of such an objective function
by deriving it as the (negative) log-posterior of an appropriate
probabilistic model. This approach naturally lead to the use
of quadratic regularization in the factors. Again, gradient descent in
the factors was used to perform the optimization.
Also, this paper introduced a logistic mapping between the low-rank matrix 
and the recorded ratings.

Recently, this line of work was pushed further in \cite{SS10},
which emphasize the advantage of using a non-uniform quadratic
regularization in the factors. The basic objective function was 
again a sum of square residuals, and version of stochastic gradient descent 
was used to optimize it.

This rich and successful line of work emphasizes  the importance
of obtaining a rigorous understanding of methods based on
local minimization of the sum of square residuals with respect to the
factors. The present paper provides a first step in that direction. Hopefully 
the techniques developed here will be useful to analyze the many variants
of this approach.

The relationship between the non-convex objective function and 
convex relaxation introduced by \cite{Fazel} was further investigated by 
\cite{SRJ05} and \cite{RFP07}. The basic relation is provided by the
identity 
\begin{eqnarray}
\|M\|_* = \frac{1}{2}\, 
\min_{M=XY^T}\big\{\|X\|^2_F+\|Y\|^2_F\big\}\, , \label{eq:Nuclear}
\end{eqnarray}
where $\|M\|_*$ denotes the nuclear norm of $M$ (the sum of its singular 
values). In other words, adding a regularization term that is quadratic in the
factors (as the one used in much of the literature reviewed above)
is equivalent to weighting $M$ by its nuclear norm, that can be regarded
as a convex surrogate of its rank.

In view of the identity (\ref{eq:Nuclear}) it might be possible to
use the results in this paper to prove stronger guarantees
on the nuclear norm minimization approach. Unfortunately this implication is
not immediate. Indeed in the present paper we assume the correct
rank $r$ is known, while on the other hand we do not use
a quadratic regularization in the factors. (See \citealp{KOImpl09} for
a procedure that estimates the rank from the data 
and is provably successful under the hypotheses of Theorem \ref{thm:main2}.) 
Trying to establish such an implication, and clarifying the
relation between the two approaches is nevertheless a 
promising research direction.
%
%
\subsection{On the Spectrum of Sparse Matrices and the Role of Trimming}
\label{sec:trimming}

The trimming step of the {\sc OptSpace} algorithm
is somewhat counter-intuitive in that we seem to be
wasting information. In this section we want to clarify 
its role through a simple example. Before 
describing the example, let us stress once again two
facts: $(i)$ In the last step of our the algorithm,
the trimmed entries are actually incorporated in the cost
function and hence the full information is exploited;
$(ii)$ Trimming is not the only way to treat over-represented
rows/columns in $M^E$, and probably not the optimal one. 
One might for instance rescale the entries of such rows/columns.
We stick to trimming because we can prove it actually works.
 
Let us now turn to the example. Assume, for the sake
of simplicity, that $m=n$, there is no noise in the revealed entries, and $M$ is the rank one
matrix with $M_{ij}=1$ for all $i$ and $j$.
Within the independent sampling model, the matrix 
$M^E$ has i.i.d. entries, with distribution 
Bernoulli$(\eps/n)$. The number of non-zero entries 
in a column is Binomial$(n,\eps/n)$ and 
is independent for different columns.
It is not hard to realize that the column with the largest number of
entries has more than $C\,\log n/\log\log n$ entries, with positive probability
(this probability can be made as large as we want  by reducing $C$).
Let $i$ be the index of this column, and
consider the test vector $\ue^{(i)}$ that has the $i$-th entry equal
to $1$ and all the others equal to $0$. By computing $\|M^E\ue^{(i)}\|$,
we conclude that the largest singular value of $M^E$ 
is at least $\sqrt{C\log n/\log\log n}$. 
In particular, this is very different from the largest 
singular value of $\E\{M^E\}=(\eps/n)M$ which is $\eps$.
This suggests that approximating $M$ with the $\T_r(M^E)$ 
leads to a large error. 
Hence trimming is crucial in proving Theorem \ref{thm:main1}. 
Also, the phenomenon is more severe in 
real data sets than in the present model, where each entry is revealed
independently.  

Trimming is also crucial in proving Theorem \ref{thm:noisy}.   
Using the above argument, 
it is possible to show that under the worst case model, 
\begin{eqnarray*}
	\|Z^E\|_2 \ge C'(\eps)\, Z_{\rm max}\, \sqrt{\frac{\log n}{\log\log n}}\; .
\end{eqnarray*}
This suggests that the largest singular value of the noise matrix $Z^E$ is 
quite different from the largest singular value of $\E\{Z^E\}$ which is 
$\eps Z_{\rm max}$.


To summarize, Theorems \ref{thm:main1} and \ref{thm:noisy}
(for the worst case model) simply do not hold without 
trimming or a similar procedure to normalize rows/columns of $N^E$.
Trimming allows to overcome the above phenomenon by 
setting to $0$ over-represented rows/columns.

%
%
\section{Proof of Theorem \ref{thm:main1}}

As explained in the introduction, the crucial idea
is to consider the singular value decomposition
of the trimmed matrix $\tN^E$ 
instead of the original matrix $N^E$.
Apart from a trivial rescaling,
these singular values are close to the ones of the original matrix $M$.
\begin{lemma}\label{lem:singularvalues}
There exists a numerical constant $C$ such that, 
with probability greater than $1-1/n^3$, 
\begin{eqnarray*}
\left|\frac{\sigma_q}{\eps}-\Sigma_q\right| \le C\Mmax\sqrt\frac{\alpha}{\eps} + \frac{1}{\eps}\|\tZ^E\|_2\;, 
\end{eqnarray*}
where it is understood that $\Sigma_q=0$ for $q>r$.
\end{lemma}
\begin{proof}
For any matrix A, let $\sigma_q(A)$ denote the $q$th singular value of $A$. 
Then, $\sigma_q(A+B)\leq\sigma_q(A)+\sigma_1(B)$, whence
\begin{eqnarray*}
\left|\frac{\sigma_q}{\eps}-\Sigma_q\right| 
 &\le& \left|\frac{\sigma_q(\tM^E)}{\eps}-\Sigma_q\right| + \frac{\sigma_1(\tZ^E)}{\eps}  \\
 &\le& C\Mmax\sqrt\frac{\alpha}{\eps} + \frac{1}{\eps}\|\tZ^E\|_2\;,  
\end{eqnarray*}
where the second inequality follows from the next Lemma 
as shown by \cite{KOM09}.

\begin{lemma}[Keshavan, Montanari, Oh, 2009]\label{lem:spectralnorm}
There exists a numerical constant $C$ such that, with probability larger than $1-1/n^3$,
\begin{eqnarray*}
\frac{1}{\sqrt{mn}}\left|\left| M-\frac{\sqrt{mn}}{\eps}\tM^E \right|\right|_2 \le C\Mmax\sqrt\frac{\alpha}{\eps} \, .
\end{eqnarray*}
\end{lemma}
\end{proof}
We will now prove Theorem \ref{thm:main1}.
\begin{proof} (Theorem \ref{thm:main1})
For any matrix $A$ of rank at most $2r$, $\|A\|_{F}\le \sqrt{2r}\|A\|_2$,
whence
\begin{eqnarray*}
\frac{1}{\sqrt{mn}}\|M-\T_r(\tN^E)\|_F 
 &\le&  \frac{\sqrt{2r}}{\sqrt{mn}}\left|\left|M - 
  \frac{\sqrt{mn}}{\eps} \Big(\tN^E-\sum_{i\geq r+1}\sigma_ix_iy_i^T\Big) \right|\right|_2 \\
 &=&  \frac{\sqrt{2r}}{\sqrt{mn}}\left|\left|M - 
  \frac{\sqrt{mn}}{\eps} \Big(\tM^E +  \tZ^E -\sum_{i\geq r+1}\sigma_ix_iy_i^T\Big) \right|\right|_2 \\
  &=&  \frac{\sqrt{2r}}{\sqrt{mn}}\left|\left| \left( M - \frac{\sqrt{mn}}{\eps} \tM^E \right) + 
  \frac{\sqrt{mn}}{\eps} \left(  \tZ^E   -\Big( \sum_{i\geq r+1}\sigma_ix_iy_i^T\Big) \right) \right|\right|_2 \\
 &\le&  \frac{\sqrt{2r}}{\sqrt{mn}}\left(\Big|\Big|M-\frac{\sqrt{mn}}{\eps}\tM^E\Big|\Big|_2 + \frac{\sqrt{mn}}{\eps}\|\tZ^E\|_2 + \frac{\sqrt{mn}}{\eps}\sigma_{r+1}\right)\\
 &\le& 2C\Mmax\sqrt\frac{2\alpha r}{\eps} \,+\, \frac{2\sqrt{2 r}}{\eps}\, \|\tZ^E\|_2\, \\
&\le& C'\Mmax\,\left(\frac{nr\alpha^{3/2}}{|E|}\right)^{1/2}\, +\, 2\sqrt{2} \left( \frac{n\sqrt{r\alpha}}{|E|}\right)\, \|\tZ^E\|_2\, .
\end{eqnarray*}
where on the fourth line, we have used the fact that for any matrices $A_i$, $\|\sum_{i}A_i \|_2 \le \sum_{i} \|A_i\|_2$.
This proves our claim.
\end{proof}

%
%
\section{Proof of Theorem \ref{thm:main2}}

Recall that the cost function is defined over the
Riemannian manifold $\Manif(m,n) \equiv\Grass(m,r)\times\Grass(n,r)$.
The proof of Theorem \ref{thm:main2} consists in controlling
the behavior of $F$ in a neighborhood of $\Um = (U,V)$ (the point 
corresponding to the matrix $M$ to be reconstructed). Throughout the proof we
let $\Co(\mu)$ be the set of matrix couples $(X,Y)
\in\reals^{m\times r}\times\reals^{n\times r}$ such that 
$\|X^{(i)}\|^2\le \mu r,\;
\|Y^{(j)}\|^2\le \mu r$ for all $i,j$.

\subsection{Preliminary Remarks and Definitions}

Given $\Xm_1=(X_1,Y_1)$ and $\Xm_2 = (X_2,Y_2)\in \Manif(m,n)$, 
two points on this manifold, their distance is defined 
as $d(\Xm_1,\Xm_2)= \sqrt{d(X_1,X_2)^2+d(Y_1,Y_2)^2}$, where,
letting $(\cos\theta_1,\dots,\cos\theta_r)$ be the singular values
of $X_1^TX_2/m$,
\begin{eqnarray*}
d(X_1,X_2) = \|\theta\|_2\, .
\end{eqnarray*}
The next remark bounds the distance between two points on the manifold. 
In particular, we will use this to bound the distance between 
the original matrix $M=U\Sigma V^T$ and the starting point of the manifold optimization 
$\hM=X_0S_0Y_0^T$.
\begin{remark}[Keshavan, Montanari, Oh, 2009]
\label{remark:Near}
Let $U,X\in\reals^{m\times r}$ with $U^TU=X^TX=m\id$,
$V,Y\in\reals^{n\times r}$ with $V^TV=Y^TY=n\id$,
and $M= U\Sigma V^T$, $\hM=XSY^T$ for $\Sigma 
= \diag(\Sigma_1,\dots,\Sigma_r)$ and $S\in\reals^{r\times r}$.
If $\Sigma_1,\dots,\Sigma_r\ge \Sigma_{\rm min}$, then
\begin{eqnarray*}
d(U,X)\le \frac{\pi}{\sqrt{2\alpha}n\Sigma_{\rm min}}\, \|M-\hM\|_F\, \;\;\; ,\;\;\;\;\;
d(V,Y)\le  \frac{\pi}{\sqrt{2\alpha}n\Sigma_{\rm min}}\, \|M-\hM\|_F
\end{eqnarray*}
\end{remark}

Given $S$ achieving the minimum in Eq.~(\ref{eq:MinimizeS}),
it is also convenient to introduce the notations
\begin{eqnarray*}
d_-(\Xm,\Um) \equiv\sqrt{\Sigma_{\rm min}^2d(\Xm,\Um)^2+\|S-\Sigma\|_F^2}\, ,\\
d_+(\Xm,\Um) \equiv\sqrt{\Sigma_{\rm max}^2d(\Xm,\Um)^2+\|S-\Sigma\|_F^2}\, .
\end{eqnarray*}
%
%
\subsection{Auxiliary Lemmas and Proof of Theorem \ref{thm:main2}}

The proof is based on the following two lemmas
that generalize and sharpen analogous bounds in \cite{KOM09}.
%
\begin{lemma}\label{lemma:Quadratic}
There exist numerical constants $C_0, C_1, C_2$
such that the following happens.
Assume $\eps\ge C_0\mu_0r\sqrt{\alpha}\,\max\{\,\log n\,;\,\mu_0r\sqrt{\alpha}(\Sigma_{\rm min}/\Sigma_{\rm max})^4\,\}$ 
and $\delta\le \Sigma_{\rm min}/(C_0\Sigma_{\rm max})$.
Then, 
\begin{eqnarray}
 F(\Xm)-F(\Um)& \ge 
  &C_1n\eps\sqrt{\alpha}\,  d_-(\Xm,\Um)^2-C_1n\sqrt{r\alpha}\|Z^E\|_2d_+(\Xm,\Um)\, ,
  \label{eq:Quadratic1}\\
 F(\Xm)-F(\Um)&\le & C_2n\eps\sqrt{\alpha}\,\Sigma_{\rm max}^2\,
  d(\Xm,\Um)^2+C_2n\sqrt{r\alpha}\|Z^E\|_2d_+(\Xm,\Um)\, ,
  \label{eq:Quadratic2}
\end{eqnarray}
for all $\Xm\in\Manif(m,n)\cap \Co(4\mu_0)$ such that $d(\Xm,\Um)\le \delta$,
with probability at least $1-1/n^4$.
Here $S\in\reals^{r\times r}$ is the matrix realizing the
minimum in Eq.~(\ref{eq:MinimizeS}).
\end{lemma}

\begin{coro}\label{coro:Quadratic}
There exist a constant $C$ such that, under the hypotheses of  Lemma 
\ref{lemma:Quadratic}
\begin{eqnarray*}
\|S-\Sigma\|_F \le C\Sigma_{\rm max}d(\Xm,\Um) + C\frac{\sqrt{r}}{\eps}\,
\|Z^E\|_2\, .
\end{eqnarray*}
Further, for an appropriate choice of the  constants in Lemma 
\ref{lemma:Quadratic}, we have
\begin{eqnarray}
\sigma_{\rm max}(S)\le 2\Sigma_{\rm max}+C \frac{\sqrt{r}}{\eps}\, 
\|Z^E\|_2\, ,\label{eq:EvalueBound1}\\
\sigma_{\rm min}(S)\ge \frac{1}{2}\Sigma_{\rm min}-
C \frac{\sqrt{r}}{\eps}\, \|Z^E\|_2\, .\label{eq:EvalueBound2}
\end{eqnarray}
\end{coro}
\begin{lemma}\label{lemma:Gradient}
There exist numerical constants $C_0, C_1, C_2$
such that the following happens. Assume 
$\eps\ge C_0\mu_0r\sqrt{\alpha}\,(\Sigma_{\rm max}/\Sigma_{\rm min})^2 \max\{\,\log n\,;\,\mu_0r\sqrt{\alpha}(\Sigma_{\rm max}/\Sigma_{\rm min})^4\,\}$ 
and $\delta\le \Sigma_{\rm min}/(C_0\Sigma_{\rm max})$.
Then, 
\begin{eqnarray}
\|\grad \tF(\Xm)\|^2 \ge C_1\, n\eps^2\, \Sigma_{\rm min}^4
\left[
d(\Xm,\Um)-C_2\frac{\sqrt{r}\Sigma_{\rm max}}{\eps\Sigma_{\rm min}}
\frac{\|Z^E\|_2}{\Sigma_{\rm min}}\right]_+^2\, ,
\label{eq:GradientLowerBound}
\end{eqnarray}
for all $\Xm\in\Manif(m,n)\cap \Co(4\mu_0)$ such that $d(\Xm,\Um)\le \delta$,
with probability at least $1-1/n^4$. (Here $[a]_+\equiv\max(a,0)$.)
\end{lemma}

We can now turn to the proof of our main theorem.
\begin{proof}(Theorem \ref{thm:main2}). 
Let $\delta= \Sigma_{\rm min}/C_0\Sigma_{\rm max}$ 
with $C_0$ large enough so that the hypotheses of Lemmas \ref{lemma:Quadratic}
and \ref{lemma:Gradient} are verified.

Call $\{\Xm_k\}_{k\ge 0}$ the sequence of pairs $(X_k,Y_k)\in \Manif(m,n)$
generated by gradient descent.
By  assumption the right-hand side of Eq.~(\ref{eq:MainBound}) 
is smaller than $\Sigma_{\rm min}$. 
The following is therefore true for some numerical constant $C$:
\begin{eqnarray}
\|Z^E\|_2\le \frac{\eps}{C\sqrt{r}}\left(\frac{\Sigma_{\rm min}}
{\Sigma_{\rm max}}\right)^2\Sigma_{\rm min}\, .\label{eq:WLOG2}
\end{eqnarray}
Notice that the constant appearing here can be made as large
as we want by modifying the constant appearing in the statement of the theorem.
Further, by using Corollary \ref{coro:Quadratic} 
in Eqs.~(\ref{eq:Quadratic1}) and (\ref{eq:Quadratic2}) we get 
\begin{eqnarray}
F(\Xm)-F(\Um)& \ge 
&C_1n\eps\sqrt{\alpha} \Sigma_{\rm min}^2 \big\{ d(\Xm,\Um)^2-\delta_{0,-}^2\big\}\, ,
\label{eq:Quadratic1New}\\
F(\Xm)-F(\Um)&\le & C_2n\eps\sqrt{\alpha}\Sigma_{\rm max}^2
\big\{d(\Xm,\Um)^2+\delta_{0,+}^2\big\}\, ,
\label{eq:Quadratic2New}
\end{eqnarray}
with $C_1$ and $C_2$ different from those in 
Eqs.~(\ref{eq:Quadratic1}) and (\ref{eq:Quadratic2}), where 
\begin{eqnarray*}
\delta_{0,-} \equiv C\frac{\sqrt{r}\Sigma_{\rm max}}{\eps
\Sigma_{\rm min}}\, \frac{\|Z^E\|_2}{\Sigma_{\rm min}}\, ,
\;\;\;\;\;\;\;\;
\delta_{0,+} \equiv C \frac{\sqrt{r}\Sigma_{\rm max}}{\eps
\Sigma_{\rm min}}\, \frac{\|Z^E\|_2}{\Sigma_{\rm max}}\, .
\end{eqnarray*}
By Eq.~(\ref{eq:WLOG2}), with large enough $C$, 
we can assume $\delta_{0,-}\leq \delta/20$ 
and $\delta_{0,+}\leq (\delta/20)(\Sigma_{\rm min}/\Sigma_{\rm max})$.

Next, we provide a bound on $d(\Um,\Xm_0)$.
Using Remark \ref{remark:Near}, we have 
$d(\Um,\Xm_0)\le (\pi/n\sqrt{\alpha}\Sigma_{\rm min})\|M-X_0S_0Y_0^T\|_F$.
Together with Theorem \ref{thm:main1} this implies 
\begin{eqnarray*}
	d(\Um,\Xm_0) \le \frac{C \Mmax\,}{\Sigma_{\rm min}} \left(\frac{r\alpha }{\eps}\right)^{1/2}\, +\,   \frac{C'\,\sqrt{r}}{\eps \Sigma_{\rm min}}\, \|\tZ^E\|_2 \;.
\end{eqnarray*}
Since $\eps\ge C''\alpha\mu_1^2r^2(\Sigma_{\rm max}/\Sigma_{\rm min})^4$ as per our assumptions  
and $\Mmax \leq \mu_1\sqrt{r}\Sigma_{\rm max}$ for incoherent $M$, 
the first term in the above bound is upper bounded by 
$\Sigma_{\rm min}/20 C_0 \Sigma_{\rm max}$, 
for large enough $C''$. 
Using Eq.~(\ref{eq:WLOG2}), with large enough constant $C$, 
the second term in the above bound is upper bounded by 
$\Sigma_{\rm min}/20 C_0 \Sigma_{\rm max}$. 
Hence we get
\begin{eqnarray*}
d(\Um,\Xm_0)\le \frac{\delta}{10}\, .
\end{eqnarray*}

We make the following claims :
\begin{enumerate}
\item $\Xm_k\in\Co(4\mu_0)$ for all $k$.

First we notice that we can assume $\Xm_0\in \Co(3\mu_0)$.
Indeed, if this does not hold, we can `rescale' those rows of $X_0$, $Y_0$ 
that violate the constraint. A proof that this
rescaling is possible was given in \cite{KOM09} (cf. Remark 6.2 there). 
We restate the result here for the reader's convenience in the
next Remark.

\begin{remark}\label{rem:rescaling}
Let $U, X \in \reals^{n \times r}$ with $U^TU = X^TX = n\id$ and $U \in \Co(\mu_0)$ and $d(X,U) \le \delta \le \frac{1}{16}$. Then there exists $X' \in \reals^{n \times r}$ such that $X'^TX' = n\id$, $X' \in \Co(3\mu_0)$ and $d(X',U) \le 4 \delta$. Further, such an $X'$ can be computed from $X$ in a time of $O(nr^2)$.
\end{remark}

Since $\Xm_0\in\Co(3\mu_0)$ ,  $\tF(\Xm_0) = F(\Xm_0) \le 
4C_2n\eps\sqrt{\alpha}\Sigma_{\rm max}^2 \delta^2/100$. 
On the other hand 
$\tF(\Xm)\ge \rho(e^{1/9}-1)$ for $\Xm\not\in\Co(4\mu_0)$.
Since $\tF(\Xm_k)$ is a non-increasing sequence, the thesis follows 
provided we take 
$\rho\ge  C_2n\eps\sqrt{\alpha}\Sigma_{\rm min}^2$.
\item  $d(\Xm_k,\Um)\le \delta/10$ for all $k$.

Since $\eps \geq C\alpha\mu_1^2r^2 (\Sigma_{\rm max}/\Sigma_{\rm min})^6 $ 
as per our assumptions in Theorem \ref{thm:main2}, we have 
$d(\Xm_0,\Um)^2 \leq (C_1\Sigma_{\rm min}^2/C_2\Sigma_{\rm max}^2)(\delta/20)^2$.
Also assuming Eq.~(\ref{eq:WLOG2}) with large enough $C$, 
we have $\delta_{0,-}\leq \delta/20$ 
and $\delta_{0,+}\leq (\delta/20)(\Sigma_{\rm min}/\Sigma_{\rm max})$.
Then, by Eq.~(\ref{eq:Quadratic2New}), 
\begin{eqnarray*}
F(\Xm_0) \leq F(\Um) + C_1n\eps\sqrt{\alpha} \Sigma_{\rm min}^2 
\,\frac{2\delta^2}{400}
\, .
\end{eqnarray*}
Also, using Eq.~(\ref{eq:Quadratic1New}), 
for all $\Xm_k$ such that $d(\Xm_k,\Um) \in [\delta/10,\delta]$, 
we have 
\begin{eqnarray*}
F(\Xm) \geq F(\Um) + C_1n\eps\sqrt{\alpha} \Sigma_{\rm min}^2 
\frac{3\delta^2}{400}\,
.
\end{eqnarray*}
Hence, for all $\Xm_k$ such that $d(\Xm_k,\Um) \in [\delta/10,\delta]$,
we have $\tF(\Xm) \geq F(\Xm) \geq F(\Xm_0)$.
This contradicts the monotonicity of $\tF(\Xm)$, and thus 
proves the claim.

\end{enumerate}

Since the cost function is twice differentiable,
and because of the above two claims, the sequence $\{\Xm_k\}$
converges to
\begin{eqnarray*}
\Omega = \big\{ \Xm\in\Co(4\mu_0)\cap\Manif(m,n)\, :\, d(\Xm,\Um)\le \delta\, ,
\grad\tF(\Xm) = 0\, \big\}\, .
\end{eqnarray*}
By Lemma \ref{lemma:Gradient} for any  $\Xm\in\Omega$,
\begin{eqnarray}
d(\Xm,\Um)\le C\frac{\sqrt{r}\Sigma_{\rm max}}{\eps\Sigma_{\rm min}}\frac{\|Z^E\|_2}{\Sigma_{\rm min}}\;. \label{eq:FinalDist}
\end{eqnarray}
Using Corollary \ref{coro:Quadratic}, we have 
$d_+(\Xm,\Um) \leq \Sigma_{\rm max} d(\Xm,\Um) + \|S-\Sigma\|_F \leq C\Sigma_{\rm max} d(\Xm,\Um) + C(\sqrt{r}/\eps)\|Z^E\|_2$. 
Together with Eqs.~(\ref{eq:BoundDiff}) and (\ref{eq:FinalDist}), this implies 
\begin{eqnarray*}
	\frac{1}{n\sqrt{\alpha}}\|M-XSY^T\|_F \leq C\frac{\sqrt{r}\Sigma_{\rm max}^2\|Z^E\|_2}{\eps\Sigma_{\rm min}^2}\;,
\end{eqnarray*}
which finishes the proof of Theorem \ref{thm:main2}. 
\end{proof}
%
%
\subsection{Proof of Lemma \ref{lemma:Quadratic} and Corollary \ref{coro:Quadratic}}

\begin{proof}(Lemma \ref{lemma:Quadratic})
The proof is based on the analogous bound in the noiseless
case, that is, Lemma 5.3 in \cite{KOM09}. 
For readers' convenience, the result is reported in Appendix \ref{app:NoiselessLemma},
Lemma \ref{lemma:QuadraticNoiseless}. For the proof of these lemmas, we refer to \cite{KOM09}.

In order to prove the lower bound, we start by noticing that 
\begin{eqnarray*}
F(\Um) \le \frac{1}{2}\|\cP_E(Z)\|^2_F\, ,
\end{eqnarray*}
which is simply proved by using $S=\Sigma$ in Eq.~(\ref{eq:MinimizeS}).
On the other hand, we have
\begin{eqnarray}
F(\Xm) & = &\frac{1}{2}\|\cP_E(XSY^T-M-Z)\|_F^2 \nonumber \\
&=& \frac{1}{2}\|\cP_E(Z)\|^2_F+ \frac{1}{2}\|\cP_E(XSY^T-M)\|_F^2-
\<\cP_E(Z),(XSY^T-M)\>\label{eq:Intermediate}\\
&\ge & F(\Um) +Cn\eps\sqrt{\alpha}\, d_-(\Xm,\Um)^2-\sqrt{2r} \|Z^E\|_2
\|XSY^T-M\|_F\, , \nonumber
\end{eqnarray}
where in the last step we used Lemma \ref{lemma:QuadraticNoiseless}.
Now by triangular inequality
\begin{eqnarray}
\|XSY^T-M\|_F^2&\le &3\|X(S-\Sigma)Y^T\|_F^2+3\|X\Sigma(Y-V)^T\|_F^2+
 3\|(X-U)\Sigma V^T\|_F^2 \nonumber\\
&\le &3nm \|S-\Sigma\|_F^2
 +3n^2\alpha\Sigma_{\rm max}^2(\frac{1}{m}\|X-U\|_F^2+\frac{1}{n}\|Y-V\|_F^2) \nonumber\\
&\le &Cn^2\alpha d_+(\Xm,\Um)^2\, , \label{eq:BoundDiff}
\end{eqnarray}

In order to prove the upper bound, we proceed as above to get
\begin{eqnarray*}
F(\Xm) & \le \frac{1}{2}\|\cP_E(Z)\|^2_F +Cn\eps\sqrt{\alpha}\Sigma_{\rm max}^2
\, d(\Xm,\Um)^2
+\sqrt{2r\alpha} \|Z^E\|_2 Cn d_+(\Xm,\Um)\, .
\end{eqnarray*}
Further, by replacing $\Xm$ with $\Um$ in Eq.~(\ref{eq:Intermediate})
\begin{eqnarray*}
F(\Um) &\ge &\frac{1}{2}\|\cP_E(Z)\|^2_F-\<\cP_E(Z),(U(S-\Sigma)V^T)\>\\
&\ge &\frac{1}{2}\|\cP_E(Z)\|^2_F-\sqrt{2r\alpha} \|Z^E\|_2 Cn d_+(\Xm,\Um)\, .
\end{eqnarray*}
By taking the difference of these inequalities we get the desired upper bound.
\end{proof}

\begin{proof}(Corollary \ref{coro:Quadratic})
By putting together Eq.~(\ref{eq:Quadratic1}) and (\ref{eq:Quadratic2}),
and using the definitions of $d_{+}(\Xm,\Um)$, $d_-(\Xm,\Um)$,
we get 
\begin{eqnarray*}
\|S-\Sigma\|_F^2 \le \frac{C_1 + C_2}{C_1}\Sigma_{\rm max}^2d(\Xm,\Um)^2 +
\frac{(C_1+C_2)\sqrt{r}}{C_1\eps}\|Z^E\|_2\sqrt{\Sigma_{\rm max}^2d(\Xm,\Um)^2
+\|S-\Sigma\|_F^2}\, .
\end{eqnarray*}
%
Let $x \equiv \|S-\Sigma\|_F$, 
$a^2 \equiv \big((C_1+C_2)/C_1\big)\Sigma_{\rm max}^2d(\Xm,\Um)^2$,
and $b \equiv \big((C_1+C_2)\sqrt{r}/C_1\eps\big)\|Z^E\|_2$. 
The above inequality then takes the form 
\begin{eqnarray*}
x^2\le a^2+b\sqrt{x^2+a^2}\le a^2+ab+bx\, ,
\end{eqnarray*}
which implies our claim $x\le a+b$.

The singular value bounds (\ref{eq:EvalueBound1}) and (\ref{eq:EvalueBound2})
follow by triangular inequality. For instance
\begin{eqnarray*}
\sigma_{\rm min}(S)\ge \Sigma_{\rm min} - C\Sigma_{\rm max}d(\Xm,\Um)-C\frac{\sqrt{r}}{\eps}\|Z^E\|_2\, .
\end{eqnarray*}
which implies the inequality  (\ref{eq:EvalueBound2})
for $d(\Xm,\Um)\le \delta = \Sigma_{\rm min}/C_0\Sigma_{\rm max}$
and $C_0$ large enough. An analogous argument proves 
Eq.~(\ref{eq:EvalueBound1}).
\end{proof}
%
%
%
\subsection{Proof of Lemma \ref{lemma:Gradient}}

Without loss of generality we will assume $\delta\le 1$,
$C_2\ge 1$ and 
\begin{eqnarray}
\frac{\sqrt{r}}{\eps}\, \|Z^E\|_2\le \Sigma_{\rm min}\, ,\label{eq:WLOG}
\end{eqnarray}
because otherwise the lower bound (\ref{eq:GradientLowerBound})
is trivial for all $d(\Xm,\Um)\le \delta$. 

Denote by $t\mapsto \Xm(t)$, $t\in[0,1]$, the geodesic 
on $\Manif(m,n)$ such that $\Xm(0) = \Um$ and $\Xm(1) =\Xm$,
parametrized proportionally to the arclength.
Let $\Whm = \dot{\Xm}(1)$ be its final velocity, with 
$\Whm = (\hW,\hQ)$. Obviously $\Whm\in\Tang_{\Xm}$ (with 
$\Tang_{\Xm}$ the tangent space of $\Manif(m,n)$ at $\Xm$)
and 
\begin{eqnarray*}
 \frac{1}{m}\|\hW\|^2 + \frac{1}{n}\|\hQ\|^2= d(\Xm,\Um)^2 , 
\end{eqnarray*}
because $t\mapsto\Xm(t)$
is parametrized proportionally to the arclength.

Explicit expressions for $\Whm$ can be obtained
in terms of $\Wm \equiv \dot{\Xm}(0)=(W,Q)$ \citep{KOM09}.
If we let $W= L\Theta R^T$ be the singular value decomposition
of $W$, we obtain
\begin{eqnarray}
\hW = -UR\Theta\sin\Theta\, R^T + L\Theta\cos\Theta\, R^T\, .\label{eq:Tangent1}
\end{eqnarray}

It was proved in \cite{KOM09} that $\<\grad G(\Xm),\Whm\>\ge 0$. 
It is therefore sufficient to lower bound the scalar product
$\<\grad F,\Whm\>$. By computing the gradient of $F$
we get
\begin{eqnarray}
\<\grad F(\Xm), \Whm\> & = & 
\<\cP_E(XSY^T-N),(XS\hQ^T+\hW SY^T)\> \nonumber \\
& =& 
\<\cP_E(XSY^T-M),(XS\hQ^T+\hW SY^T)\>-
\<\cP_E(Z),(XS\hQ^T+\hW SY^T)\> \nonumber \\
& = & \<\grad F_0(\Xm), \Whm\>-\<\cP_E(Z),(XS\hQ^T+\hW SY^T)\>\,
 \label{eq:GradF}
\end{eqnarray}
where $F_0(\Xm)$ is the cost function in absence of noise,
namely 
\begin{eqnarray}
F_0(X,Y) = \min_{S\in\reals^{r\times r}}
\left\{\frac{1}{2}\sum_{(i,j)\in E}
\big((XSY^T)_{ij}-M_{ij}\big)^2\right\}\, .\label{eq:F0}
\end{eqnarray}
As proved in \cite{KOM09}, 
\begin{eqnarray}
\<\grad F_0(\Xm),\Whm\>\ge Cn\epsilon\sqrt{\alpha}\Sigma_{\rm min}^2d(\Xm,\Um)^2\, 
\label{eq:GradF0}
\end{eqnarray}
%
(see Lemma \ref{lemma:GradientNoiselessBis} in Appendix).

We are therefore left with the task of upper bounding
$\<\cP_E(Z),(XS\hQ^T+\hW SY^T)\>$. Since $XS\hQ^T$ has rank 
at most $r$, we have
\begin{eqnarray*}
\<\cP_E(Z),XS\hQ^T\>\le \sqrt{r}\, \|Z^E\|_2\, \|XS\hQ^T\|_F\, .
\end{eqnarray*}
Since $X^TX = m\id$, we get
\begin{eqnarray}
\|XS\hQ^T\|^2_F & = &m\Trace(S^TS\hQ^T\hQ)\le 
n\alpha \sigma_{\rm max}(S)^2 \|\hQ\|_F^2 \nonumber \\
&\le& Cn^2\alpha\Big(\Sigma_{\rm max}+\frac{\sqrt{r}}{\eps}\|Z^E\|_F\Big)^2
\, d(\Xm,\Um)^2 \label{eq:XSZtbound}\\
&\le& 4Cn^2\alpha\Sigma_{\rm max}^2
\, d(\Xm,\Um)^2\,  , \nonumber
\end{eqnarray}
where, in inequality (\ref{eq:XSZtbound}), we used Corollary \ref{coro:Quadratic} 
and in the last step, we used Eq.~(\ref{eq:WLOG}). 
Proceeding analogously for $\<\cP_E(Z),\hW SY^T\>$, we get 
\begin{eqnarray*}
\<\cP_E(Z),(XS\hQ^T+\hW SY^T)\>
\le C'n\Sigma_{\rm max}\sqrt{r\alpha}\, \|Z^E\|_2\, d(\Xm,\Um)\, .
\end{eqnarray*}
Together with Eq.~(\ref{eq:GradF}) and (\ref{eq:GradF0}) this implies
\begin{eqnarray*}
\<\grad F(\Xm),\Whm\>\ge C_1n\epsilon\sqrt{\alpha}\Sigma_{\rm min}^2d(\Xm,\Um)
\Big\{d(\Xm,\Um)
-C_2\frac{\sqrt{r}\Sigma_{\rm max}}{\eps\Sigma_{\rm min}}
\frac{\|Z^E\|_2}{\Sigma_{\rm min}}\Big\}\, ,
\end{eqnarray*}
which implies Eq.~(\ref{eq:GradientLowerBound}) by Cauchy-Schwartz inequality.

%
%

\section{Proof of Theorem \ref{thm:noisy}}

\begin{proof}({\em Independent entries model })
We start with a claim that for any sampling set $E$, we have 
$$\|\tZ^E\|_2 \leq \|Z^E\|_2\;.$$ 
To prove this claim, let $x^*$ and $y^*$ be 
$m$ and $n$ dimensional vectors, respectively, 
achieving the optimum in $\max_{\|x\|\le 1,\|y\|\le 1}\{x^T\tZ^E y\}$,  
that is, such that $\|\tZ^E\|_2 = x^{*T} \tZ^E y^*$.
Recall that, as a result of the trimming step, 
all the entries in trimmed rows and columns of $\tZ^E$ are set to zero. 
Then, there is no gain in maximizing $x^T\tZ^E y$ 
to have a non-zero entry $x^*_i$  
for $i$ corresponding to the rows which are trimmed. 
Analogously, for $j$ corresponding to the trimmed columns, 
we can assume without loss of generality that $y^*_j=0$. 
From this observation, it follows that $x^{*T} \tZ^E y^* = x^{*T} Z^E y^*$, 
since the trimmed matrix $\tZ^E$ and the sample noise matrix $Z^E$ only 
differ in the trimmed rows and columns.
The claim follows from the fact that $x^{*T} Z^E y^* \leq \|Z^E\|_2$, 
for any $x^*$ and $y^*$ with unit norm.

In what follows, we will first prove that $\|Z^E\|_2$ is bounded by 
the right-hand side of Eq.~(\ref{eq:boundindepcase}) for any range of $|E|$. 
Due to the above observation, this implies that $\|\tZ^E\|_2$ is 
also bounded by $C\sigma \sqrt{\eps\sqrt{\alpha}\log n}$,
where $\eps\equiv |E|/\sqrt{\alpha}n$.
Further, we use the same analysis to prove a tighter bound 
in Eq.~(\ref{eq:boundindepcase2}) when $|E|\ge n\log n$. 

First, we want to show that $\|Z^E\|_2$ is 
bounded by $C\sigma \sqrt{\eps\sqrt{\alpha}\log n}$, 
 and 
$Z_{ij}$'s are i.i.d. random variables 
with zero mean and sub-Gaussian tail with parameter $\sigma^2$.
The proof strategy is to show that $\E\big[\|Z^E\|_2\big]$ is bounded, using 
the result of \cite{Seginer00} on expected norm of random matrices, 
and use the fact that $\|\,\cdot\, \|_2$ is a Lipschitz continuous function of its arguments 
together with concentration inequality for Lipschitz functions on i.i.d. Gaussian random variables 
due to \cite{T96}. 

Note that $\|\cdot\|_2$ is a Lipschitz function 
with a Lipschitz constant $1$.
Indeed, for any $M$ and $M'$, 
$\big|\|M'\|_2 - \|M\|_2 \big| \leq \|M'-M\|_2 \leq \|M'-M\|_F$, 
where the first inequality follows from triangular inequality 
and the second inequality follows from the fact that 
$\|\cdot\|_F^2$ is the sum of the squared singular values. 

To bound the probability of large deviation, 
we use the result on concentration inequality for 
Lipschitz functions on i.i.d. sub-Gaussian random variables due to \cite{T96}. 
For a $1$-Lipschitz function $\|\cdot\|_2$ on $m\times n$ i.i.d. random variables $Z^E_{ij}$ 
with zero mean, and sub-Gaussian tails with parameter $\sigma^2$, 
\begin{eqnarray}
	\prob\big(  \|Z^E\|_2-\E[\|Z^E\|_2] > t \big) \leq \exp\Big\{-\frac{t^2}{2\sigma^2}\Big\}\;. \label{eq:noisytalagrand}
\end{eqnarray}
Setting $t=\sqrt{8\sigma^2\log n}$, this implies that 
$\|Z^E\|_2 \leq \E\big[\|Z\|_2\big] + \sqrt{8\sigma^2\log n}$ 
with probability larger than $1-1/n^4$.

Now, we are left to bound the expectation $\E\big[\|Z^E\|_2\big]$. 
First, we symmetrize the possibly asymmetric random variables $Z^E_{ij}$ 
to use the result of \cite{Seginer00} on expected norm of random matrices 
with symmetric random variables. 
Let $Z'_{ij}$'s be independent copies of $Z_{ij}$'s, 
and $\xi_{ij}$'s be independent Bernoulli random variables such that 
$\xi_{ij}=+1$ with probability $1/2$ and 
$\xi_{ij}=-1$ with probability $1/2$. 
Then, by convexity of $\E\big[\|Z^E-Z'^E\|_2|Z'^E\big]$ and Jensen's inequality, 
\begin{eqnarray*}
	\E\big[\|Z^E\|_2\big] \leq \E\big[\|Z^E-Z'^E\|_2\big] = \E\big[\|(\xi_{ij}(Z^E_{ij}-Z'^E_{ij}))\|_2\big] \leq 2\E\big[\|(\xi_{ij}Z^E_{ij})\|_2\big]\;,
\end{eqnarray*}
where $(\xi_{ij}Z^E_{ij})$ denotes an $m\times n$ matrix 
with entry $\xi_{ij}Z^E_{ij}$ in position $(i,j)$.
Thus, it is enough to show that $\E\big[\|Z^E\|_2\big]$ 
is bounded by $C\sigma \sqrt{\eps\sqrt{\alpha}\log n}$ 
in the case of symmetric random variables $Z_{ij}$'s.

To this end, we apply the following bound on 
expected norm of random matrices with i.i.d. 
symmetric random entries, proved by 
\citet[Theorem 1.1]{Seginer00}.
%
	\begin{eqnarray}
		\E\big[\|Z^E\|_2\big] \leq C \Big( \E\big[\max_{i\in[m]}{\|Z^E_{i\bullet}\|}\big] + \E\big[\max_{j\in[n]}{\|Z^E_{\bullet j}\|\big]} \Big)\;,
	\label{eq:expectednorm}
	\end{eqnarray}
	where $Z^E_{i\bullet}$ and $Z^E_{\bullet j}$ denote the $i$th row and $j$th column of $A$ respectively.
For any positive parameter $\beta$, which will be specified later, the following is true.
\begin{eqnarray}
	\E\big[\max_j\|Z^E_{\bullet j}\|^2\big] \leq \beta\sigma^2\eps\sqrt{\alpha} + \int_{0}^{\infty}{ \prob\big(\max_j\|Z^E_{\bullet j}\|^2 \geq \beta\sigma^2\eps\sqrt{\alpha} + z \big) \, \de z } \,. \label{eq:boundexpectednorm}
\end{eqnarray}
To bound the second term, we can apply union bound on 
each of the $n$ columns, and use 
the following bound on each column $\|Z^E_{\bullet j}\|^2$ resulting from concentration of measure inequality 
for the i.i.d. sub-Gaussian random matrix $Z$.
\begin{eqnarray}
	\prob\Big( \sum_{k=1}^{m} (Z^E_{kj})^2 \geq \beta\sigma^2\eps\sqrt{\alpha} + z \Big) \leq 
\exp\Big\{-\frac{3}{8}\Big((\beta-3)\eps\sqrt{\alpha} + \frac{z}{\sigma^2}\Big)\Big\}\;. \label{eq:boundchernoff}
\end{eqnarray}

To prove the above result, we apply Chernoff bound on the sum of independent random variables. 
Recall that $Z^E_{kj}=\txi_{kj}Z_{kj}$ where $\txi$'s are independent Bernoulli random variables 
such that $\txi=1$ with probability $\eps/\sqrt{mn}$ and zero with probability $1-\eps/\sqrt{mn}$. Then, 
for the choice of $\lambda = 3/8\sigma^2 < 1/2\sigma^2$, 
\begin{eqnarray*}
	\E\Big[ \exp\Big( \lambda \sum_{k=1}^{m} (\txi_{kj}Z_{kj})^2 \Big) \Big] 
		& = &\Big(1-\frac{\eps}{\sqrt{mn}} + \frac{\eps}{\sqrt{mn}}\E[e^{\lambda Z_{kj}^2}] \Big)^m \\
		& \leq & \Big(1-\frac{\eps}{\sqrt{mn}} + \frac{\eps}{\sqrt{mn(1-2\sigma^2\lambda)}} \Big)^m \\
		& = & \exp\Big\{ m\log\Big(1+\frac{\eps}{\sqrt{mn}}\Big) \Big\} \\
		& \leq & \exp\big\{ \eps\sqrt{\alpha} \big\} \;,
\end{eqnarray*}
where the first inequality follows from the definition of 
$Z_{kj}$ as a zero mean random variable with sub-Gaussian tail, and 
the second inequality follows from $\log(1+x)\leq x$.
By applying Chernoff bound, Eq.~(\ref{eq:boundchernoff}) follows. 
Note that an analogous result holds for the Euclidean norm on 
the rows $\|Z^E_{i\bullet}\|^2$.

Substituting Eq.~(\ref{eq:boundchernoff}) and 
$\prob\big(\max_j\|Z^E_{\bullet j}\|^2 \geq z \big) \leq m\,\prob\big(\|Z^E_{\bullet j}\|^2 \geq z \big)$ in Eq.~(\ref{eq:boundexpectednorm}), we get 
\begin{eqnarray}
	\E\big[\max_j\|Z^E_{\bullet j}\|^2\big] \leq \beta\sigma^2\eps\sqrt{\alpha} + \frac{8\sigma^2m}{3}e^{-\frac{3}{8}(\beta-3)\eps\sqrt{\alpha}}\;.
\label{eq:boundexpectednormbeta}
\end{eqnarray}
The second term can be made arbitrarily small by taking $\beta=C\log n$ with large enough $C$.
Since $\E\big[\max_j\|Z^E_{\bullet j}\|\big] \leq \sqrt{\E\big[\max_j\|Z^E_{\bullet j}\|^2\big]}$, 
applying Eq.~(\ref{eq:boundexpectednormbeta}) with 
$\beta=C\log n$ in Eq.~(\ref{eq:expectednorm}) gives 
\begin{eqnarray*}
	\E\big[\|Z^E\|_2\big] \leq C\sigma\sqrt{\eps\sqrt{\alpha}\log n}\;.
\end{eqnarray*}
Together with Eq.~(\ref{eq:noisytalagrand}), 
this proves the desired thesis for any sample size $|E|$. 

In the case when $|E|\ge n\log n$, 
we can get a tighter bound by similar analysis. 
Since $\eps\geq C' \log n$, for some constant $C'$, 
the  second term in Eq.~(\ref{eq:boundexpectednormbeta}) 
can be made arbitrarily small with a large constant $\beta$.
Hence, applying Eq.~(\ref{eq:boundexpectednormbeta}) 
with $\beta=C$ in Eq.~(\ref{eq:expectednorm}), we get
\begin{eqnarray*}
	\E\big[\|Z^E\|_2\big] \leq C\sigma\sqrt{\eps\sqrt{\alpha}}\;.
\end{eqnarray*}
Together with Eq.~(\ref{eq:noisytalagrand}), 
this proves the desired thesis for $|E|\ge n\log n$. 

%

\end{proof}

\begin{proof}({\em Worst Case Model })
 Let $D$ be the $m \times n$ all-ones matrix.
Then for any matrix $Z$ from the \emph{worst case model}, 
we have $\|\tZ^E\|_2\leq Z_{\rm max}\|\tD^E\|_2$, 
since $x^T\tZ^Ey \leq \sum_{i,j} Z_{\rm max}|x_i|\tD^E_{ij}|y_j|$,
which follows from the fact that $Z_{ij}$'s are uniformly bounded.
Further, $\tD^E$ is an adjacency matrix of a corresponding 
bipartite graph with bounded degrees.
Then, for any choice of $E$ the following is true for all positive
integers $k$: 
\begin{eqnarray*}
 \|\tD^E\|_2^{2k} \leq \max_{x,\|x\|=1}\big|x^T((\tD^E)^T\tD^E)^k x\big| 
 \leq \text{Tr}\big( ((\tD^E)^T \tD^E)^k \big) \leq n(2\eps)^{2k}\;.
\end{eqnarray*}
Now  $\text{Tr}\big( ((\tD^E)^T \tD^E)^k \big)$
is the number of paths of length $2k$ on the bipartite graph with 
adjacency matrix $\tD^E$,  that begin and end at $i$ for every $i \in [n]$.
Since this graph has degree bounded by $2\eps$, we get
\begin{eqnarray*}
 \|\tD^E\|_2^{2k} \leq n(2\eps)^{2k}\;.
\end{eqnarray*}
Taking $k$ large, we get the desired thesis.
\end{proof}

%
%
%
\acks{This work was partially supported by
a Terman fellowship, the NSF CAREER award CCF-0743978
and the NSF grant DMS-0806211. 
SO was supported by a fellowship from the Samsung Scholarship Foundation. }

%
%
\appendix 
%
%
\section{Three Lemmas on the Noiseless Problem}
\label{app:NoiselessLemma}
\begin{lemma}\label{lemma:QuadraticNoiseless}
There exists numerical constants $C_0, C_1, C_2$
such that the following happens. Assume 
$\eps\ge C_0 \mu_0r\sqrt{\alpha}\,\max\{\,\log n\,;\,\mu_0r\sqrt{\alpha}(\Sigma_{\rm min}/\Sigma_{\rm max})^4\,\}$ 
and $\delta\le \Sigma_{\rm min}/(C_0\Sigma_{\rm max})$.
Then, 
\begin{eqnarray*}
 C_1 \sqrt{\alpha}\,\Sigma_{\rm min}^2\, d(\Xm,\Um)^2+C_1\sqrt{\alpha}\, \|S_0-\Sigma\|_{F}^2
\le \frac{1}{n\eps}\, F_0(\Xm)\le  C_2\sqrt{\alpha}\Sigma_{\rm max}^2
d(\Xm,\Um)^2\,,
\end{eqnarray*}
for all $\Xm\in\Manif(m,n)\cap \Co(4\mu_0)$ such that $d(\Xm,\Um)\le \delta$,
with probability at least $1-1/n^4$.
Here $S_0\in\reals^{r\times r}$ is the matrix realizing the
minimum in Eq.~(\ref{eq:F0}).
\end{lemma}

\begin{lemma}\label{lemma:GradientNoiseless}
There exists numerical constants $C_0$ and $C$
such that the following happens.
Assume 
$\eps\ge C_0 \mu_0r\sqrt{\alpha}\,(\Sigma_{\rm max}/\Sigma_{\rm min})^2 \max\{\,\log n \,;\,\mu_0r\sqrt{\alpha}(\Sigma_{\rm max}/\Sigma_{\rm min})^4\,\} $ 
and $\delta\le \Sigma_{min}/(C_0\Sigma_{\rm max})$.
Then 
\begin{eqnarray*}
\|\grad \tF_0(\Xm)\|^2 \ge C\, n\eps^2\,\Sigma_{\rm min}^4 d(\Xm,\Um)^2\,,
\end{eqnarray*}
for all $\Xm\in\Manif(m,n)\cap \Co(4\mu_0)$ such that $d(\Xm,\Um)\le \delta$,
with probability at least $1-1/n^4$.
\end{lemma}

\begin{lemma}\label{lemma:GradientNoiselessBis}
Define $\Whm$ as in Eq.~(\ref{eq:Tangent1}). 
Then there exists numerical constants $C_0$ and $C$
such that the following happens.
Under the hypothesis of Lemma \ref{lemma:GradientNoiseless} 
\begin{eqnarray*}
\<\grad F_0(\Xm),\Whm\> \ge C\, n\eps\sqrt{\alpha}\,\Sigma_{\rm min}^2 d(\Xm,\Um)^2\,,
\end{eqnarray*}
for all $\Xm\in\Manif(m,n)\cap \Co(4\mu_0)$ such that $d(\Xm,\Um)\le \delta$,
with probability at least $1-1/n^4$.
\end{lemma}



\bibliography{MatrixCompletion}

\end{document}